# New Encoder Learning for Captioning Heavy Rain Images via Semantic Visual Feature Matching

*Chang-Hwan Son and Pung-Hwi Ye

Department of Software Convergence Engineering, Kunsan National University

558 Daehak-ro, Gunsan-si 54150, Republic of Korea

*Corresponding Author

Phone Number: 82-63-469-8915; Fax Number: 82-63-469-7432

E-MAIL: changhwan76.son@gmail.com; cson@kunsan.ac.kr

## Abstract

Image captioning generates text that describes scenes from input images. It has been developed for high-quality images taken in clear weather. However, in bad weather conditions, such as heavy rain, snow, and dense fog, the poor visibility owing to rain streaks, rain accumulation, and snowflakes causes a serious degradation of image quality. This hinders the extraction of useful visual features and results in deteriorated image captioning performance. To address practical issues, this study introduces a new encoder for captioning heavy rain images. The central idea is to transform output features extracted from heavy rain input images into semantic visual features associated with words and sentence context. To achieve this, a target encoder is initially trained in an encoder–decoder framework to associate visual features with semantic words. Subsequently, the objects in a heavy rain image are rendered visible by using an initial reconstruction subnetwork (IRS) based on a heavy rain model. The IRS is then combined with another semantic visual feature matching subnetwork (SVFMS) to match the output features of the IRS with the semantic visual features of the pretrained target encoder. The proposed encoder is based on the joint learning of the IRS and SVFMS. It is trained in an end-to-end manner, and then connected to the pretrained decoder for image captioning. It is experimentally demonstrated that the proposed encoder can generate semantic visual features associated with words even from heavy rain images, thereby increasing the accuracy of the generated captions.

## 1. Introduction

Image captioning [1] is a technique that can describe a scene from an input image, including the behavior of objects and their interactions. Specifically, image captioning generates captions from input images. This requires a high-level visual understanding of the objects and their interactions in the image. It also requires an understanding of word embedding [2] to transform words into meaningful vectors, as well as a statistical language model that can sequentially arrange words corresponding to object attributes and behaviors according to a probability distribution over word sequences. Therefore, image captioning is a multimodal fusion [3] involving both image and text processing. This is considerably more challenging than traditional computer vision problems such as object detection [4] and image classification [5].

### 1.1. Motivation

With the advent of deep learning, image captioning has made significant progress. However, existing image captioning methods [1,6-8] target high-quality images captured in clear weather, and few techniques have been developed for low-quality images captured in bad weather conditions, such as heavy rain, snow, and deep fog [9-12]. In such conditions, image artifacts, such as rain streaks, noise, blurring, and snowflakes, can appear in such images. In particular, serious visibility degradation occurs because of rain accumulation [13], which is due to rain streak accumulation in the line of sight, and it forms a layer of veil in the background. This rain accumulation results in washed-out images and slightly blurred distant-background scenes. Low visibility and image artifacts hinder the extraction of useful visual features from low-quality images and cause deterioration of image captioning performance. In this study, we are only concerned with images captured in heavy rain conditions.

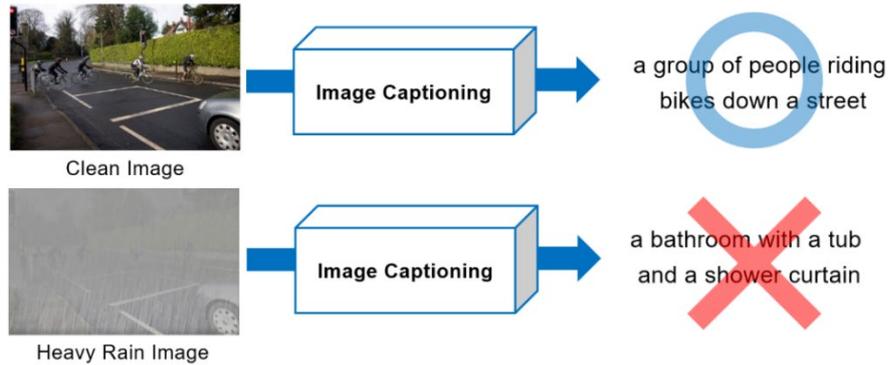

Fig. 1. Necessity of the development of the proposed image captioning method.

Fig. 1 shows an example indicating the difficulty of captioning heavy rain images. Unlike the clean images targeted by existing image captioning methods, heavy rain images (considered in this study) may exhibit serious degradation; specifically, low visibility caused by rain accumulation and image artifacts such as rain streaks and slight blurring, as shown in the bottom left of Fig. 1. This results in poor image captioning performance. For example, the rain streaks in the figure cause the image captioning method in [8] to generate incorrect descriptions such as "a shower curtain". Therefore, it is necessary to address these image degradation issues for more accurate captioning of heavy rain images.

## 1.2. Proposed Approach

In this study, we develop a new image captioning method that can produce semantic visual features associated with words and generate accurate captions even for heavy rain images. To the best of our knowledge, this study is the first attempt in this direction. Fig. 2 compares conventional image captioning methods [1,8] to the proposed approach. Conventional image captioning methods consist of two main modules. One is the encoder and the other is the decoder. Additionally, there may be another attention module (not shown in the figure for clarity). Encoders generate compact visual codes from clean images using pre-trained models, and decoders predict the current word from previous words using the encoded visual codes. However, in heavy rain conditions, severe visibility degradation occurs in captured images as a result of rain accumulation, which prevents conventional encoders from extracting accurate visual codes from heavy rain images. Therefore, a new encoder should be designed to address the image degradation problem and extract accurate visual features from heavy rain images. In the proposed approach, a conventional decoder is used.

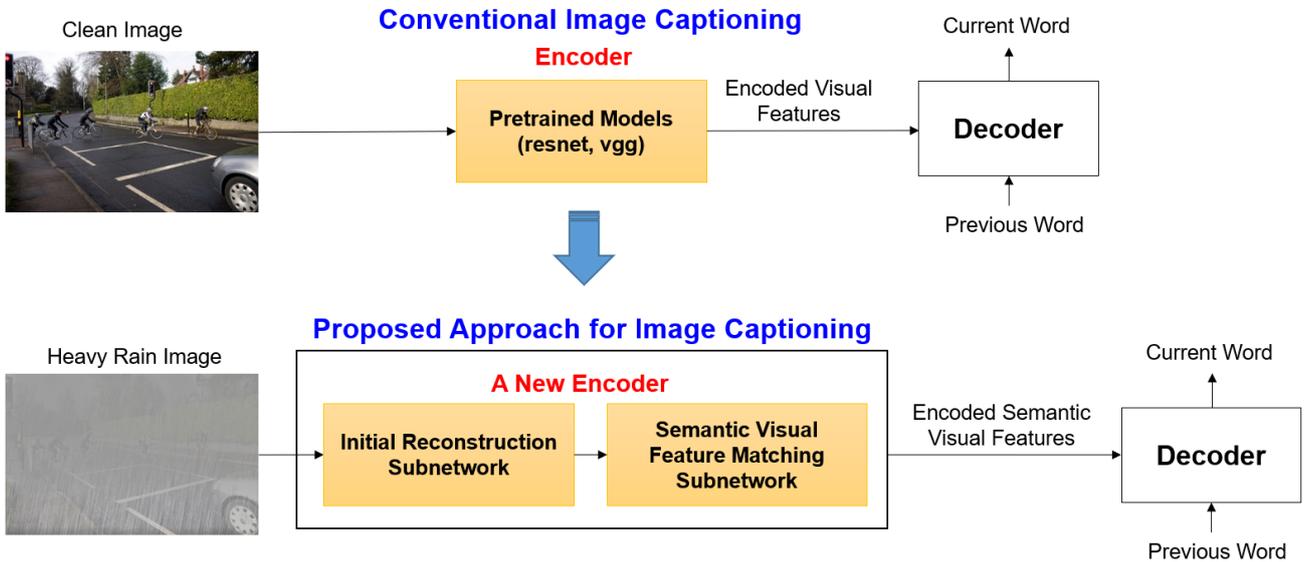

Fig. 2. Conventional image captioning versus the proposed approach.

The proposed encoder consists of two subnetworks. One is the initial reconstruction subnetwork (IRS) and the other is the semantic visual feature matching subnetwork (SVFMS). To encode visual information from low-quality heavy rain images, the IRS first eliminates artifacts and improves visibility. However, the ultimate goal of the proposed encoder is not rain removal, but the extraction of semantic visual features. Therefore, the SVFMS is used to transform the output features extracted by the IRS into semantic visual features associated with words. The IRS serves as a guide that aids the SVFMS in finding semantic visual features by defining an appropriate initial solution. Semantic visual features refer to the output features of the target encoder, which was pre-trained jointly with the decoder to consider the sentence context and extract semantic visual features from clean images. By training the proposed encoder, which is based on the joint training of the IRS and SVFMS, our goal can be achieved and semantic visual features associated with words can be extracted from heavy rain images, thereby improving the generated captions.

### 1.3. Contributions

- Unlike in the case of conventional neural-network-based image captioning [1,6-8], this study is concerned with practical image degradation issues caused by heavy rain conditions. To this end, we introduce a new encoder architecture so that semantic visual features associated with words can be extracted from heavy rain images. To the best of our knowledge, this is the first attempt in this direction. In particular, we incorporate rain removal and semantic visual feature matching into the proposed encoder. Moreover, it is experimentally demonstrated that the proposed encoder can extract semantic visual features from heavy rain images and improve the generated captions. Accordingly, the proposed encoder may be used as a new base model for image captioning in heavy rain conditions.

- This study provides new training and test datasets for captioning heavy rain images. Even though there are a few image captioning datasets, such as MSCOCO [14] and Flickr30 [15], they contain only clean images captured in clear weather or indoors. Therefore, to reflect heavy rain conditions, such as rain streaks, blurring, and rain accumulation, a new image dataset is required. That is, heavy rain images based on the heavy rain model [13] are required. The heavy rain images used in this study, along with their captions, will be publicly available for research purposes. This new dataset may be used as a reference dataset for captioning heavy rain images. The evaluation scores may also be used for performance comparisons.

## 2. Related Works

This study deals with practical image captioning problems that occur in heavy rain conditions and presents a new encoder architecture that incorporates rain removal and semantic visual feature matching into the encoder-decoder framework. Therefore, this section introduces related works on rain removal and image captioning.

### 2.1. Rain removal

Single-image rain removal methods [16] are categorized into model-based and data-driven methods. Model-based approaches employ optimization frameworks consisting of a data fidelity term and prior term. The data-fidelity term measures the accuracy of rain synthesis models such as additive [16] and nonlinear composite models [13], which can predict real rain images. The prior term models handcrafted priors regarding rain shape and direction. Sparse coding [16–18] and Gaussian mixture models [19] were widely used prior models before the development of deep learning methods. Data-driven approaches use a large amount of observed data and automatically extract rich and hierarchical features through a layer-by-layer transformation. That is, data-driven approaches are deep-learning techniques. Since the introduction of the detail network [20], various architectures, such as density-aware [21], joint rain detection and removal [13], scale-aware [22,23], and progressive networks [24], have been developed. However, no research has been conducted on the effectiveness of these rain removal methods as a preprocessing step in computer vision applications, such as object detection and image captioning for heavy rain images.

### 2.2. Image captioning

Earlier approaches to image captioning are based on templates [25]. Specifically, they generate caption templates, the slots of which are filled in according to the output of object detection, attribute classification, and scene recognition. Recent approaches [1] are based on neural networks and use an encoder–decoder framework; they are inspired by the success of sequence-to-sequence learning applied to machine translation. In the encoder–decoder framework, a convolutional neural network (CNN) and a recurrent neural network (RNN) are used for encoding and decoding, respectively. However, the CNN does not allow the encoded visual features to adapt to the sentence context, and the RNN cannot model long-range dependencies because of the vanishing and exploding gradient problems. To address these issues, the RNN is replaced by long short-term memory (LSTM) [8] for decoding, and an attention mechanism is incorporated into the encoder–decoder framework. The attention mechanism models the spatial importance of each visual feature to evolve adaptively to the sentence context. More advanced attention mechanisms, such as channel attention [26], visual sentinel [27], and attention on attention [28] have been developed. However, these methods target high-quality clean images captured in clear weather. As no research has been conducted on image captioning methods that consider bad weather conditions, there is a need to address practical issues, such as low visibility and image artifacts, that can occur in images captured in bad weather conditions.

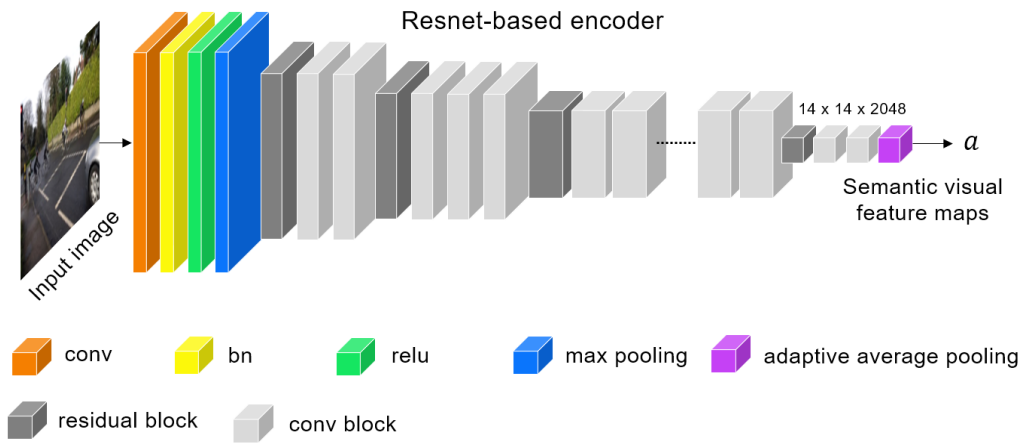

Fig. 3. Encoder architecture.

## 3. Backgrounds

Herein, we introduce a basic image captioning method that is based on the widely used encoder–decoder framework with attention [8]. In particular, conventional encoders are described in detail to highlight their differences from the proposed encoder.

### 3.1. Encoder

Encoders extract visual features from input images. In encoding, high-dimensional input images are mapped to low-dimensional visual feature vectors. Pretrained models, such as vgg and resnet, have become the basis of encoders. The architecture of a resnet-based encoder for image captioning is shown in Fig. 3. The resnet-based encoder differs from resnet in that the last fully connected layers are removed, and an adaptive average pooling layer is added to fix the size of the output feature maps. The other network components, that is, **conv**olution (**conv**), **b**atch **n**ormalization (**bn**), **re**ctified **l**inear **u**nit (**relu**), max pooling, residual block, and conv block, are the same as in resnet. Further details can be found in [29,8]. The resnet-based encoder finally outputs a three-dimensional (3D) tensor through a layer-by-layer feature transformation:

$$\boldsymbol{a} = \{\boldsymbol{a}_1, \boldsymbol{a}_2, .., \boldsymbol{a}_L\} = f_{encoder}(\boldsymbol{J}), \tag{1}$$

where $f_{encoder}$ denotes the encoder function, $\boldsymbol{J}$ denotes the clean input image, and $\boldsymbol{a}$ is the output of the encoder function; its size is $14 \times 14 \times 2048$. Subscript $i$ denotes spatial pixel location. For example, $\boldsymbol{a}_i$ indicates the $i$-th feature vector with a size of $1 \times 1 \times 2048$. The pretrained feature vectors, $\boldsymbol{a}_i$, with a largescale image dataset can carry visual information to distinguish objects. That is, the resnet-based encoder has high discriminative power, and thus it is effective in object identification.

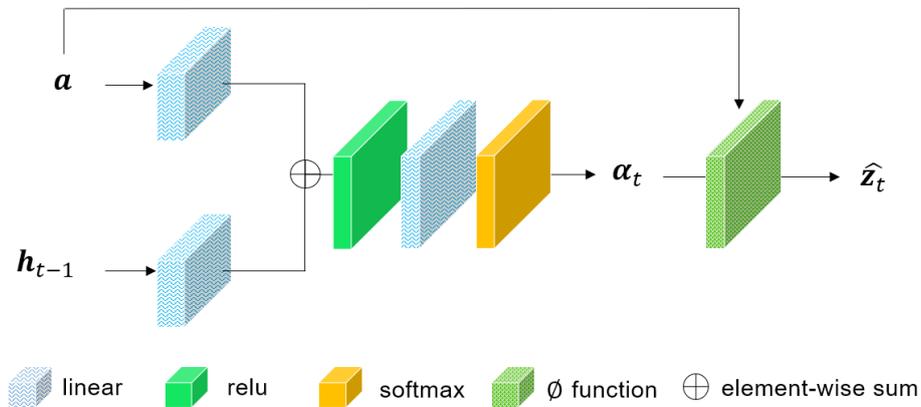

Fig. 4. Architecture of the attention module for encoding [8].

### 3.2. Attention module and decoder

In Fig. 2, the decoder uses LSTM to learn long-term dependencies. The LSTM-based decoder predicts the current from the previous word at time step $t$ using the visual features provided by the encoder. At time step $t$, the decoder employs an attention mechanism to determine spatially important visual feature vectors when predicting the current word. That is, the attention mechanism models the spatial importance of each visual feature vector $\boldsymbol{a}_i$ and produces a weighted average over the feature vectors for encoding. The architecture of the attention module is shown in Fig. 4 [8]. The attention module receives the visual feature vectors $\boldsymbol{a}$ and hidden vector $\boldsymbol{h}_{t-1}$ at time step $t$, which are given by the CNN-based encoder and LSTM, respectively. Through the linear, relu, and softmax layers, the attention module outputs a spatial map $\boldsymbol{\alpha}_t$ and then determines the final visual feature vector $\hat{\boldsymbol{z}}_t$ using a function $\boldsymbol{\phi}$:

$$\hat{\boldsymbol{z}}_t = \phi(\boldsymbol{a}, \boldsymbol{\alpha}_t) = \sum_{i=1}^{L} \boldsymbol{\alpha}_{t,i} \boldsymbol{a}_i \, , \qquad (2)$$

Here, $\boldsymbol{\phi}$ is defined as a weighted sum. The above equation indicates that only one visual feature vector is given by the attention module. This vector is also called a context vector in [26] because it is associated with semantic words and sentence context. In Eq. (2), it is noted that the hidden vector $\boldsymbol{h}_{t-1}$, which stores long-term visual and linguistic information, is also used to estimate the context vector, as shown in Fig. 4. In Eq. (2), all vectors are not constant but parameters to be learned.

Given the previous word and the context vector, the decoder predicts the current word. The previous word is first transformed into a meaningful vector using word embedding techniques such as word2vec [2] and glove [30]; subsequently, it is fed into the LSTM. In addition, both the hidden vector and memory cell stored in the LSTM are also used in decoding. The output layer of the decoder computes the word probability given the LSTM state, the context vector, and the previous word:

$$\boldsymbol{y}_t = f_{decoder}\big(\boldsymbol{E}\boldsymbol{y}_{t-1}, \hat{\boldsymbol{Z}}_t, \boldsymbol{h}_t\big) \, , \qquad (3)$$

where $\boldsymbol{E}$ denotes the word embedding matrix, $\boldsymbol{y}_{t-1}$ and $\boldsymbol{y}_t$ denote the previous and the current word, respectively, and $f_{decoder}$ is the decoder function, which outputs the final prediction $\boldsymbol{y}_t$ based on the LSTM. As we are primarily concerned with the encoder, we do not discuss other details regarding the decoder. Such details can be found in [8].

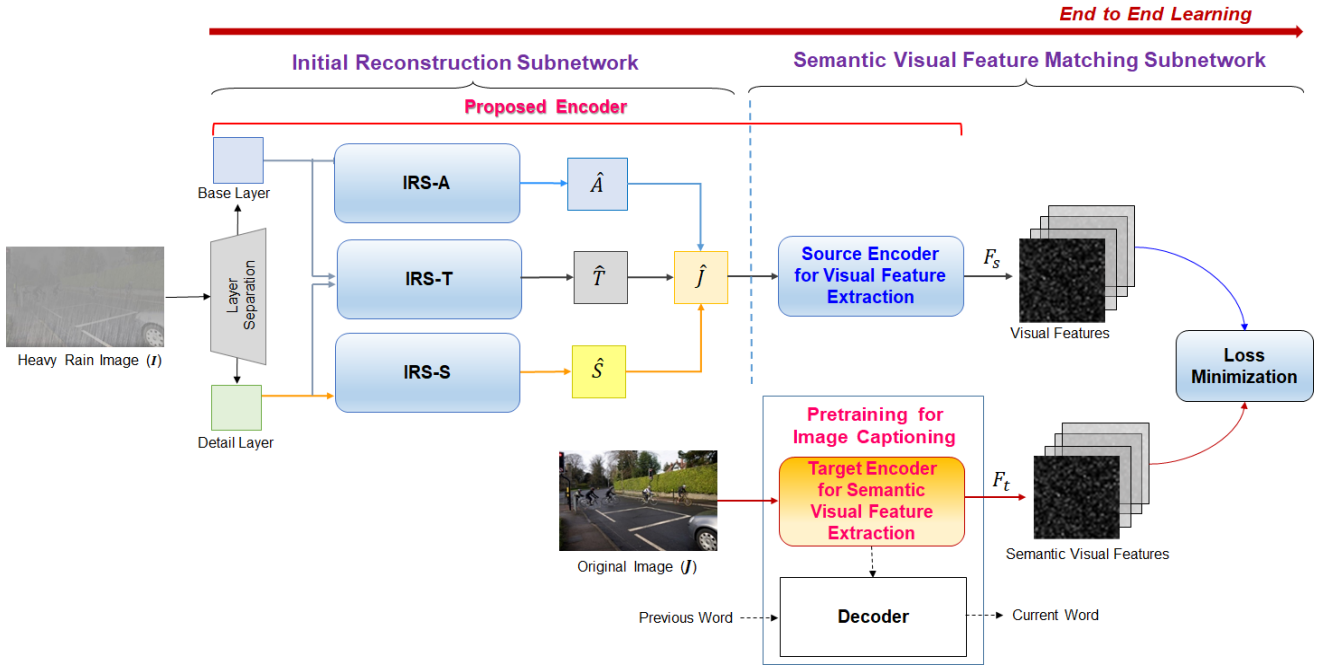

Fig. 5. Architecture of the proposed encoder for semantic visual feature matching.

## 4. Proposed Encoder

This study presents a new image captioning encoder that addresses practical degradation issues in heavy rain images, such as low-visibility and rain streaks, which prevent the extraction of useful visual features. The proposed image captioning method follows the widely used encoder-decoder framework [8] introduced in the background section. However, conventional encoders cannot handle image degradation problems caused by heavy rain. Therefore, in this study, a novel encoder is designed for image captioning under heavy rain conditions. Unlike a conventional encoder that only uses pre-trained models such as resnet or vgg for feature extraction, the proposed encoder combines rain removal and semantic visual feature matching. Its architecture is shown in Fig. 5 and comprises two subnetworks: IRS and SVFMS. The former generates initial restored images, and the latter transforms the visual features extracted from the IRS into semantic visual features. The combination of these subnetworks can be used as a new encoder for captioning heavy rain images more effectively. Details regarding the IRS and SVFMS are provided below.

## 4.1. Heavy rain model

Before presenting the IRS in detail, we discuss the modeling of heavy rain images. Unlike the additive composite model [16–19], which considers only rain streaks, the heavy-rain model [13] can describe rain accumulation, which forms a layer of veil in the background and produces washed-out images. Heavy rain images have particularly low visibility, as shown in Fig. 2, resulting in degraded image captioning performance. Therefore, we selected the heavy rain model rather than the additive composite model.

The heavy rain model has the following mathematical form [13]:

$$I = T \odot (J + \sum_{i=1}^{n} S_i) + (1 - T) \odot A, \qquad (4)$$

where $I$ denotes a heavy rain image, $T$, $J$, and $A$ denote the transmission map, original clean image, and atmospheric light map, respectively, $S_i$ is the rain layer that includes only the $i$-th rain streaks, $1$ is a matrix of ones, and $\odot$ denotes elementwise multiplication. To understand this equation, we decompose it as follows:

$$R = J + \sum_{i=1}^{n} S_i, \qquad (5)$$

$$I = T \odot R + (1 - T) \odot A, \qquad (6)$$

Here, $R$ is the rain streak image, which is the sum of the original image and the rain layers. That is, Eq. (5) corresponds to the additive composite model widely used for rain removal. However, this model cannot reflect heavy rain conditions. Accordingly, rain accumulation should be modeled. This can be achieved by using Eq.

(6), which indicates that the heavy rain image is the weighted sum of the rain streak image and the atmospheric light map.

## 4.2. Heavy rain image dataset for rain removal and image captioning

We used the popular MSCOCO2014 [14] dataset for rain removal and image captioning. This dataset contains approximately 120,000 high-quality clean images and five captions for each image, but no heavy rain images. Thus, we generated a new heavy rain image dataset. According to the heavy rain model in Eq. (4), depth maps [31] are first estimated from the clean images in the MSCOCO2014 dataset, and then $T$, $S$, and $A$ are generated to produce heavy rain images. In Eq. (4), $S_i$ is synthesized by generating Gaussian noise and applying a motion filter. The atmospheric light map $A$ is filled with the same bright pixel values, and $T$ is derived from the depth map based on the physical haze model in which the scene radiance is attenuated exponentially with depth [9]. A random number generator is used to determine the noise level, direction, and length of the motion filter, and the pixel value of atmospheric light.

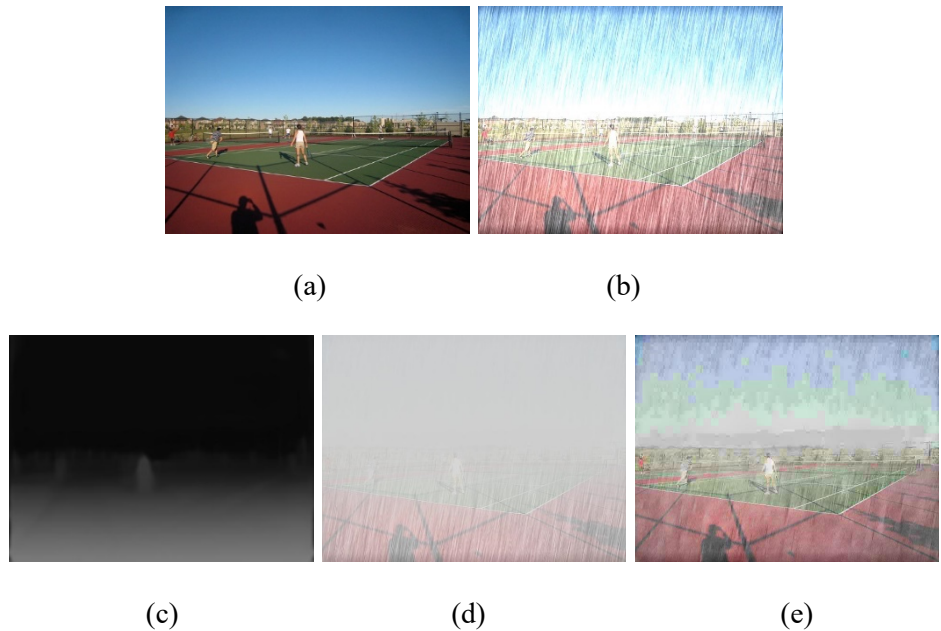

(a)                                          (b)

(c)                           (d)                           (e)

Fig. 6. Heavy rain model : (a) original image, (b) rain streak image, (c) depth map, (d) heavy rain image, and (e) image restored by the IRS

The images obtained by applying the heavy rain model are shown in Fig. 6. Specifically, the original clean image is shown in Fig. 6(a), and the rain streak image generated using Eq. (5) is shown in Fig. 6(b), where it is seen that rain streaks are overlaid on the original image. The estimated depth map, which is converted into the transmission map, is shown in Fig. 6(c). The heavy rain image generated using Eq. (4) is shown in Fig. 6(d). Compared with the rain streak image in Fig. 6(b), it exhibits particularly low visibility owing to rain accumulation. In addition, as depth increases, visibility decreases significantly.

### 4.3. Initial Reconstruction Subnetwork

The heavy rain model has unknown parameters $\boldsymbol{T}$, $\boldsymbol{S}$, and $\boldsymbol{A}$, as shown in Eq. (4). To estimate these parameters, the IRS has three subnetworks, namely, **IRS_A**, **IRS_T**, and **IRS_S**, as shown in Fig. 5. These subnetworks may have different architectures, but we use a model with an architecture similar to that of the generator in [32]. Although more complex and larger rain removal networks can be considered, our ultimate purpose is not to propose a new rain-removal network; thus, a simple and powerful architecture is used.

Parameters $\boldsymbol{T}$, $\boldsymbol{S}$, and $\boldsymbol{A}$ correspond to different image characteristics. For example, $\boldsymbol{A}$ represents atmospheric light, and it is well known that illumination has low-frequency characteristics [33]. By contrast, $\boldsymbol{S}$ represents rain structures, and thus it has high-frequency characteristics. Therefore, it is more effective to estimate $\boldsymbol{T}$, $\boldsymbol{S}$, and $\boldsymbol{A}$ by decomposing the image into a base and a detail layer, as shown in Fig. 5. Specifically, we use guided image filtering [34] to generate the base layer, whereas the detail layer is obtained by subtracting the base layer from the heavy rain image. The base and detail layers are used to estimate $\boldsymbol{A}$ and $\boldsymbol{S}$, respectively. Both layers are used to estimate $\boldsymbol{T}$:

$$L_{IRS} = \left\| f_{IRS\_A}(\boldsymbol{I}_b) - \boldsymbol{A} \right\|_2^2 + \left\| f_{IRS\_T}(\boldsymbol{I}_b, \boldsymbol{I}_d) - \boldsymbol{T} \right\|_2^2 + \left\| f_{IRS\_S}(\boldsymbol{I}_d) - \boldsymbol{S} \right\|_2^2, \tag{7}$$

$L_{IRS}$ indicates the loss to be minimize here and $\|\cdot\|_2$ denotes the $l_2$-norm. $f_{IRS\_A}$, $f_{IRS\_T}$, and $f_{IRS\_s}$ indicate the three subnetwork functions for **IRS_A**, **IRS_T**, and **IRS_S**. respectively. $\boldsymbol{I}_b$ and $\boldsymbol{I}_d$ denote the base and detail layers, respectively. For IRS training, 8,000 heavy rain images are selected from our dataset and the extracted patch is of size 128x128. The batch size are 4. The epoch number is 300 and Adam optimizer [35] is used.

After the functions have been learned, the derained image $\hat{J}$ can be obtained by reversing the process in Eq. (4), that is,

$$\hat{J} = f_{inv}(\hat{T}, \hat{A}, \hat{S}, I) = \frac{I - (1 - \hat{T}) \odot \hat{A}}{\hat{T}} - \hat{S} , \qquad (8)$$

where $\hat{T}$, $\hat{S}$, and $\hat{A}$ are the estimated transmission map, rain layer, and atmospheric light map, respectively, and $f_{inv}$ denotes the inverse function used to derive the reconstructed image.

The restored image obtained using the learned IRS is shown in Fig 6(e). It can be seen that image quality is poor, particularly for the sky regions; however, object visibility (e.g., people) is improved. Image quality is poor because the visibility of heavy rain input images is extremely low. That is, this study addresses a highly challenging problem, and as our purpose is not rain removal but captioning of heavy rain images, an excessively large and complex architecture is not considered for the IRS. For image captioning, other networks such as the SVFMS and decoder are also required. Therefore, it is inefficient to design a larger and more complex IRS architecture.

### 4.4. Semantic Visual Feature Matching Subnetwork

In Fig. 5, the SVFMS has two encoders for semantic visual feature matching: One is a target encoder, and the other is a source encoder. First, the pretrained target encoder, at the bottom of Fig. 5, was jointly trained with the decoder for image captioning. We used open source code [8] to train a widely used encoder–decoder model with attention for image captioning. The MSCOCO2014 dataset, consisting of approximately 12,0000 clean images and five captions for each image was used. Here, 110,000 clean images were used as a training set, and 5,000 clean images were used as a validation set. The remaining 5,000 clean images were used as a test set. More information on the parameter setting, loss function, and optimizer can be found in [8]. It should be noted

arXiv:2105.13753 [cs.CV]

that clean images were used for training. The pretrained target encoder outputs *semantic visual features*, which are used as a reference for semantic visual feature matching. Unlike the pretrained vgg and resnet with only image datasets, the target encoder was learned jointly with the decoder, implying that the output features of the pretrained target encoder are associated with semantic words.

Second, another source encoder is added to the back of the IRS to output features from the heavy rain input image. Therefore, these features contain only visual information, and they are indeed *visual features*, which are different from the semantic visual features output by the pretrained target encoder. Hence, the source encoder is additionally required to transform the features of the pretrained IRS into visual features. The source and target encoders have the same architecture, as shown in Fig. 3.

The SVFMS transforms the output of the pretrained IRS into semantic visual features. Therefore, the purpose of the pretrained IRS is not to remove rain streaks and rain accumulation, but to set an appropriate initial point for semantic feature extraction. A network can be designed to directly convert a heavy rain image into semantic visual features; however, this approach may be inefficient and may not easily yield an optimal solution. Therefore, the pretrained IRS serves as a guide that aids in finding semantic visual features. An initial reconstructed image is output from the pretrained IRS, and subsequently it becomes semantic visual features through semantic visual feature matching. In Fig. 5, the IRS and the source encoder appear separated for clarity, but the IRS can eventually be regarded as part of the source encoder.

## 4.5 Semantic Visual Feature Matching

Given the proposed encoder and the pretrained target encoder, the entire network in Fig. 5 is trained in an end-to-end manner for semantic visual feature matching. Here, the proposed encoder is considered a network that combines the IRS and the source encoder. It directly outputs the final visual features from the input heavy rain image. The purpose of semantic visual feature matching is to match the visual features of the proposed encoder with the semantic visual features of the pretrained target encoder. In the training phase, our dataset, that is, 80,000 pairs of heavy rain and clean images were used, and the $l_1$-norm was used to measure the prediction errors (it should be noted that the entire network requires both heavy rain images and the corresponding clean images):

$$\hat{J} = f_{inv}(f_{IRS\_A}(I_b), f_{IRS\_S}(I_d), f_{IRS\_T}(I_b, I_d), I) ,$$ (9)

$$F_S = f_{encode\_s}(\hat{J}) \quad \text{and} \quad F_T = f_{encoder\_t}(J) ,$$ (10)

$$L_{SVFM} = \|F_S - F_T\|_1 ,$$ (11)

where $L_{SVFM}$ denotes the loss to be minimized, $\|\cdot\|_1$ denotes the $l_1$-norm, and $F_S$ and $F_T$ are the visual features of the proposed encoder and the semantic visual features of the pretrained target encoder, respectively. By minimizing this loss, the visual features $F_S$ output from the proposed encoder can be close to the semantic visual features $F_T$. In Eq. (11), $F_T$ is the ground truth generated by inputting clean images into the target encoder. $f_{encoder\_t}$ is fixed, and only the parameters of the $f_{encode\_s}$ and $f_{IRS\_A,S,T}$ are learned. The Adam optimizer [35] was used to update the parameters of the entire network.

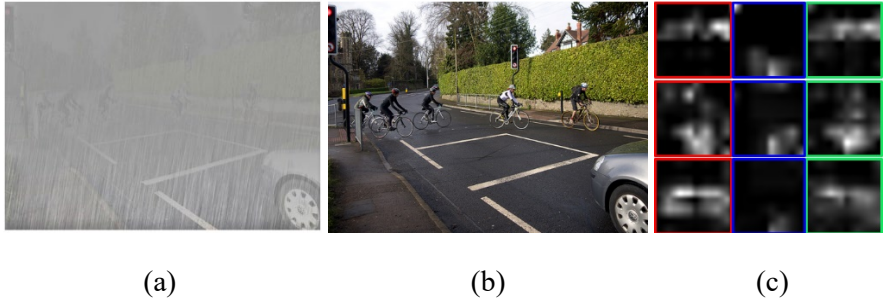

(a)                      (b)                      (c)

Fig. 7. Semantic visual feature matching: (a) heavy rain image, (b) original clean image, and (c) semantic visual feature comparison

The effectiveness of the proposed encoder when applied to a heavy rain image is clearly demonstrated in Fig. 7. Figs. 7(a) and 7(b) show heavy rain and clean images, respectively. In Fig. 7(c), the red boxes indicate the semantic visual features of the clean image passing through the target encoder. That is, these features are the targets ($\boldsymbol{F}_T$). The blue and green boxes indicate the visual features of the heavy rain image in Fig. 7(a) that are fed into the target encoder and the proposed encoder, respectively. By comparing the features in the blue and green boxes with those in the red boxes, it can be concluded that the visual features of the heavy rain image passing through the proposed encoder can be close to the semantic visual features of the clean image. That is, the proposed encoder can improve visual features from heavy rain images with particularly low visibility and rain streaks.

## 5. Experimental Results

### 5.1. Experimental settings

In the test phase, the proposed encoder, consisting of the IRS and the source encoder, is used to generate semantic visual features from heavy rain input images. After the entire network shown in Fig. 5 has been trained, the pretrained target encoder is no longer required. We used our dataset for performance evaluation. As mentioned in subsections 4.2 and 4.4, our dataset is an updated version of the MSCOCO2014 dataset, with heavy rain images added. Specifically, 5,000 pairs of heavy rain and clean images with their captions were used as a test set. Other images and captions were already used as training and validation sets for semantic visual feature matching and target encoder pretraining. The architectures were implemented using Pytorch on a Windows PC. Our source codes and datasets can be downloaded at https://github.com/cvmllab.

### 5.2. Selected framework and compared encoders

The widely used encoder–decoder framework in [8] (described in Section 2) was selected to evaluate the image captioning performance of the proposed architecture. Specifically, we fixed the pretrained decoder and changed only the encoder in the encoder–decoder framework. The same pretrained decoder and attention module were used, implying that the proposed encoder could operate in this framework. Specifically, in the proposed method, the semantic visual features output from the proposed encoder are fed into the pretrained decoder.

To evaluate the performance of conventional encoders, two different encoders were tested. One is the pretrained target encoder, which takes heavy rain or derained images as input, and then feeds the output into the decoder. The other is a pretrained source encoder, which can directly transform heavy rain input images into semantic visual features. This encoder was trained without the IRS in Fig. 5. Hereafter, the neural-network-based image captioning (NIC) method using the pretrained target and the source encoder is referred to as NIC_T and NIC_S, respectively. The only difference between the proposed encoder and NIC_S is whether the IRS is used. We compare NIC_T and NIC_S with the proposed encoder to evaluate the effectiveness of the IRS and SVFMS. NIC_T can take heavy rain or derained images as input. The latter case is referred to as NIC_T(D).

## 5.3 Qualitative evaluation

The captioning results for heavy rain images are shown in Fig. 8, where the left and middle columns correspond to the original clean and heavy rain images, respectively. It can be seen that the heavy rain images exhibit severe deterioration, and thus our task is quite challenging. The right column shows a comparison of the reference sentences with the captions generated using NIC_T, NIC_S, NIC_T(D), and the proposed encoder. The reference sentences are the captions generated by the encoder–decoder-based NIC [8], where clean images are used as inputs. For performance evaluation, only the encoder is changed, whereas the decoder is fixed. Therefore, the reference sentences can be considered the ground truth for caption comparison.

In the first row, it can be seen that the conventional NIC_T, NIC_S, and NIC_T(D) methods generate incorrect words and phrases. For example, NIC_T and NIC_S generate incorrect phrases, namely, "cat on a bed" and "a bird sitting," respectively. In addition, NIC_T(D) generates an incorrect word, namely, "baseball". NIC_T takes a heavy rain image as input and feeds it to the pretrained target encoder. Because the input is not a clean image, low visibility and rain streaks hinder the extraction of useful visual features of objects, resulting in incorrect captions. This indicates the effectiveness of the proposed semantic visual feature matching. Moreover, NIC_S directly transforms input heavy rain images into semantic visual features through the pretrained source encoder, which matches the output features of the source encoder with the semantic visual features of the target encoder. Therefore, NIC_S can find visual features associated with the word "tennis" and generate the correct expression "tennis court"; however, its performance is poor owing to image degradation. This indicates the effectiveness of the IRS. NIC_T(D) is input images derained by IRS, and rain removal improves the generated caption, namely, it can detect the man and generate the correct phrase "a man standing." This indicates the effectiveness of the IRS. However, the visibility of heavy rain images is quite low, and thus effective deraining is nontrivial. In contrast, the proposed method can detect objects such as people and the tennis court in the heavy rain image, and thus, a more accurate caption is generated. This is because the IRS can improve object visibility, and the SVMFS enables the production of semantic visual features associated with words.

| 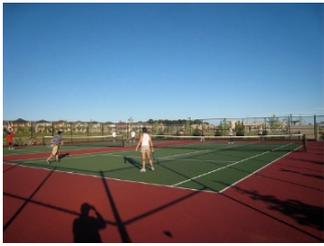 | 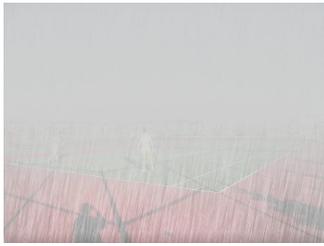 | Reference Sentence | a couple of people playing a game of tennis |
| | | NIC_T | a picture of a **cat on a bed** |
| | | NIC_S | **a bird sitting** on top of a tennis court |
| | | NIC_T(D) | a man standing on top of **a baseball field** |
| | | **Proposed Encoder** | **a couple of people playing a game of tennis** |
| 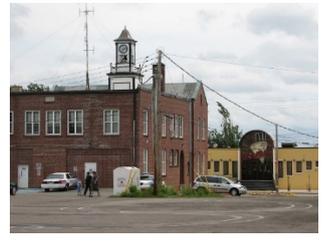 | 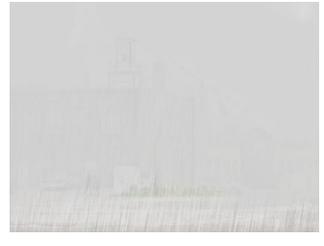 | Reference Sentence | a building with a clock tower on top of it |
| | | NIC_T | a black and white photo of **a cat** |
| | | NIC_S | a couple of **white birds** sitting on top of a building |
| | | NIC_T(D) | a black and white photo of a building with a **train** in the background |
| | | **Proposed Encoder** | **a couple of people walking down a street** |
| 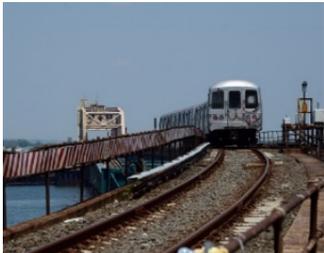 | 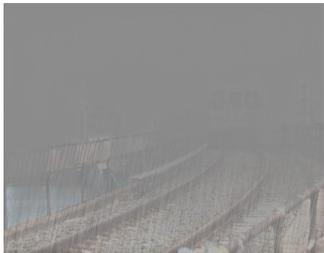 | Reference Sentence | a train traveling down train tracks next to a bridge |
| | | NIC_T | a train traveling down a track **in the snow** |
| | | NIC_S | a close up of **a person on a couch** |
| | | NIC_T(D) | **a train that is** on a train track |
| | | **Proposed Encoder** | **a train traveling down train tracks next to a bridge** |
| 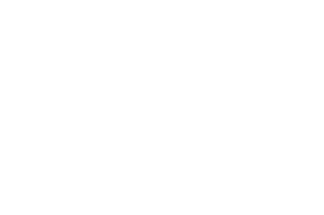 | 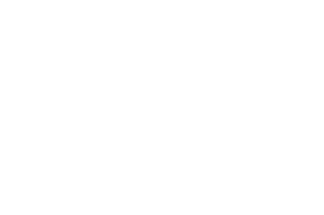 | Reference Sentence | a man playing tennis on a clay court |
| | | NIC_T | **a man riding skis** down a snow covered slope |
| | | NIC_S | **a man riding a skateboard** in a skate park |
| | | NIC_T(D) | **a man riding a wave** on top of a surfboard |
| 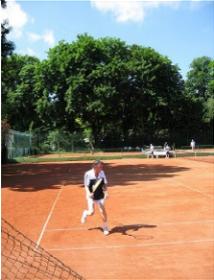 | 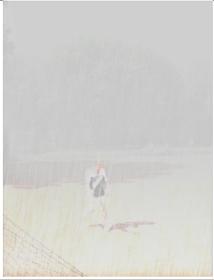 | **Proposed Encoder** | **a man is swinging a tennis racket at a ball** |
| 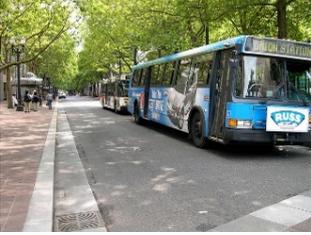 | 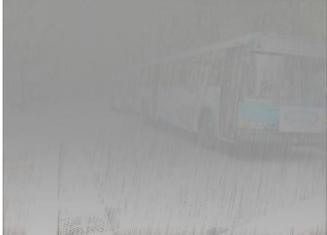 | Reference Sentence | a public transit bus on a city street |
| | | NIC_T | a picture of **a person on a wall** |
| | | NIC_S | **a close up of a person** on a skateboard |
| | | NIC_T(D) | **a bus that is sitting** in the street |
| | | **Proposed Encoder** | **a public transit bus on a city street** |
| 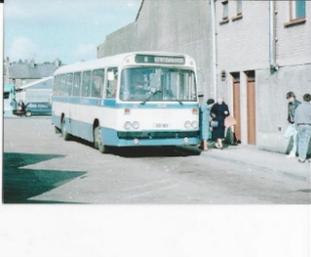 | 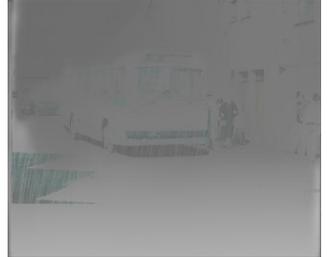 | Reference Sentence | a bus parked on the side of a street |
| | | NIC_T | a picture of a person **in the snow** |
| | | NIC_S | a close up of a person **holding a remote control** |
| | | NIC_T(D) | a group of people **standing in a room** |
| | | **Proposed Encoder** | **a bus parked on the side of the road** |
| 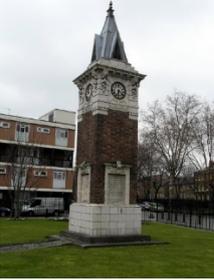 | 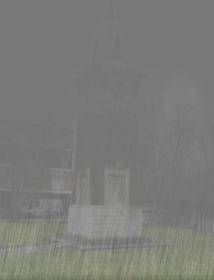 | Reference Sentence | a tall tower with a clock on top |
| | | NIC_T | a black and **white photo** of a building |
| | | NIC_S | a close up of a black and **white tie** |
| | | NIC_T(D) | **a large building** with a clock on it |
| | | **Proposed Encoder** | **a clock tower in the middle of a park** |
| 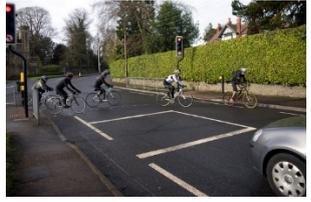 | 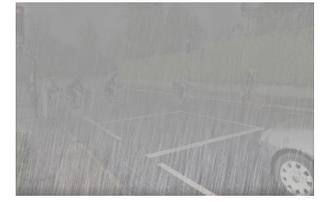 | Reference Sentence | a group of people riding bikes down a street |
| | | NIC_T | a bathroom with a tub and a **shower curtain** |
| | | NIC_S | a black and **white photo** of a car |
| | | NIC_T(D) | a person riding **a skate board** on a skate park |
| | | **Proposed Encoder** | **a group of people riding bikes down a street** |

Fig. 8. Experimental results; original clean images (left), heavy rain images (middle), and captions (right).

Similar results were obtained for other heavy rain images. In the fifth row, NIC_T and NIC_S generate incorrect phrases, namely, "a person on a wall" and "a close up of a person". NIC_T(D) can provide the improved caption "a bus that is sitting"; however, the sentence is unnatural. In contrast, the proposed encoder can extract the semantic visual features corresponding to the words "bus" and "street", and thus a more accurate caption can be generated. In the last row, the NIC_T, NIC_S, and NIC_T(D) generate incorrect phrases, namely, "shower curtain," "white photo," and "a skate board," whereas the proposed method can generate a correct phrase, namely "people riding bikes." From these results, it is confirmed that the proposed encoder can extract the semantic visual features associated with words from low-quality heavy rain images, thereby improving the generated captions.

Table 1. Quantitative evaluation of different encoders

| Metrics / Methods | Input Image Type | BLEU-1 | BLEU-2 | BLEU-3 | BLEU-4 | METEOR | ROUGE | CIDEr |
|---|---|---|---|---|---|---|---|---|
| NIC_T | Heavy Rain Images | 52.84 | 32.40 | 19.97 | 12.87 | 29.39 | 37.81 | 0.2715 |
| NIC_S | Heavy Rain Images | 54.99 | 36.10 | 24.34 | 16.67 | 32.24 | 43.65 | 0.5407 |
| NIC_T(D) | Derained Images | 61.63 | 43.22 | 30.81 | 21.98 | 38.59 | 45.93 | 0.6049 |
| **Proposed Encoder** | **Heavy Rain Images** | **68.31** | **50.87** | **37.98** | **28.13** | **44.72** | **50.47** | **0.8425** |
| NIC_T [8] (Image captioning [8] in clear weather) | Clean Images | 73.25 | 56.64 | 43.68 | 33.28 | 49.94 | 54.50 | 1.0418 |

## 5.4 Quantitative evaluation

We used BLEU[36], METEOR[37], ROUGE[38], and CIDER[39] to evaluate image captioning. The answers provided by MSCOCO2014 were used as ground truth. Table 1 shows the results of the four encoder types when the pretrained decoder and attention module were fixed. It can be seen that NIC_T has the lowest scores in all evaluations. This is because the target encoder was trained using original clean images. Therefore, NIC_T cannot extract the correct visual features. NIC_S can obtain higher scores than NIC_T. This is because NIC_S was trained to directly transform heavy rain input images into semantic visual features through the source encoder. NIC_T(D) used the derained images as input. Therefore, the visibility of the input image could be improved, leading to improved scores. However, owing to the particularly low visibility, this method cannot easily generate accurate captions. The proposed encoder outperforms the conventional NIC_T, NIC_S, and NIC_T(D) methods owing to the jointly trained IRS and SVFMS. The IRS sets an appropriate initial solution for semantic feature extraction. That is, the pretrained IRS aids in finding semantic visual features. An initial restored image is output from the pretrained IRS, and subsequently it becomes semantic visual features through the SVFMS. These results demonstrate that through the joint learning of the IRS and SVFMS, semantic visual features associated with words can be extracted from heavy rain images, thereby improving the accuracy of the generated captions.

The last row of Table 1 shows the evaluation scores of NIC_T for the clean images (i.e., images captured in clear weather). A comparison of the evaluation scores in the first row with those in the last row indicates the effect of heavy rain conditions on image captioning performance, which is reduced by approximately 20% in BLEU-1. In addition, the evaluation scores in the last row of Table 1 correspond to the maximum scores that the proposed encoder can reach. Although the proposed encoder achieved a significant improvement over NIC_T, NIC_S, and NIC_T(D), there is room for improvement.

## 5.5 Discussion and future work

In our experiments, synthetic heavy rain images were tested and the proposed encoder achieved satisfactory performance. However, it is unclear if the proposed encoder would work well on real heavy rain images. Therefore, we tested real heavy rain images downloaded from several websites. Fig. 9 presents example image captioning results. For these two images, the proposed encoder can generate relatively accurate captions. Additionally, the proposed encoder outperformed the conventional encoders.

However, for other real heavy rain images, the proposed encoder failed to generate accurate captions. This is primarily because the heavy rain model used in this study cannot accurately model real heavy rain. Specifically, the heavy rain model is incomplete, so there are visual differences between real heavy rain images and synthetic heavy rain images. This problem is referred to as *domain shift* [40]. Even though some parameters such as motion, noise level, and transmission map can be varied to generate heavy rain images for training, these synthetic heavy rain images are visually different from real heavy rain images. This domain shift problem inevitably reduces image captioning performance. However, the heavy rain model has gradually improved. As the heavy rain model becomes more accurate, it is expected that the proposed encoder will work well for real heavy rain images. Another solution is to apply a domain adaptation technique [40], which can reduce significant data distribution gaps between training and target domains. In the future, we plan to collect real heavy rain images and upgrade the proposed image captioning method using the domain adaptation technique.

| 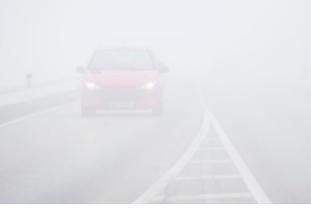 | NIC_T | <span style="color:red">a bottle of water</span> sitting on top of a sink |
| | NIC_S | a close up of a <span style="color:red">black and white</span> background |
| | NIC_T(D) | a red car driving down a <span style="color:red">train track</span> |
| | **Proposed Encoder** | **a red car driving down a street** |
| 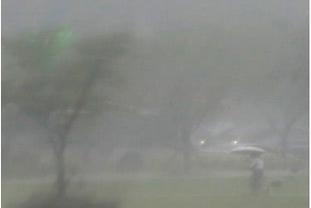 | NIC_T | a picture of a <span style="color:red">clock</span> on a cloudy day |
| | NIC_S | a close up of a person on a <span style="color:red">skateboard</span> |
| | NIC_T(D) | a person is throwing a <span style="color:red">frisbee</span> in a park |
| | **Proposed Encoder** | **a man standing on a lush green field** |

Fig. 9. Experimental results for real rain heavy images.

**Conclusion**

Unlike clean images targeted by existing image captioning methods, heavy rain images exhibit serious degradation, such as particularly low visibility and rain streaks. To address this, a new encoder based on the joint learning of the IRS and SVFMS was proposed to extract semantic visual features from heavy rain images. The principle is to transform the visual features extracted from input images into semantic visual features associated with words. To this end, a target encoder was first trained with the decoder to associate semantic words with visual features, and then the joint learning of the IRS and SVFMS enables the matching of the output features of the proposed encoder with the semantic visual features of the pretrained target encoder. It was experimentally confirmed that by using the IRS and SVFMS, semantic visual features associated with words can be extracted from low-quality heavy-rain images, thereby improving the generated captions. The proposed encoder may become a new base model for captioning heavy rain images. Moreover, the proposed architecture can be used for self-driving, in surveillance cameras, and to aid visually impaired people in heavy rain conditions. In addition, our dataset may become a reference dataset for captioning heavy rain images.

## Acknowledgment

This work was supported by the National Research Foundation of Korea(NRF) grant funded by the Korea government(MSIT) (No. 2020R1A2C1010405).